\documentclass[sigconf]{acmart}
\usepackage{tabularx,booktabs} 
\usepackage{vcell}
\usepackage{amsthm}
\usepackage{makecell}
\usepackage{enumitem}
\usepackage{multirow}
\usepackage{caption}
\newtheorem{proposition}{Proposition}

\AtBeginDocument{%
  }

\setcopyright{acmlicensed}
\copyrightyear{2025}
\acmYear{2025}
\acmDOI{XXXXXXX.XXXXXXX}
\acmConference[CIKM '25]{The 34th ACM International Conference on Information and Knowledge Management}{November 10--14,
  2025}{Coex, Seoul, Korea}
\acmISBN{978-1-4503-XXXX-X/2025/11}

\begin{document}

\title{Personalized Interpolation: Achieving Efficient Conversion Estimation with Flexible Optimization Windows}


\author{Xin Zhang}
\email{xinzhang5@meta.com}
\affiliation{%
  \institution{Meta Platforms, Inc.}
  \country{USA}
}

\author{Weiliang Li}
\email{weiliangli@meta.com}
\affiliation{%
  \institution{Meta Platforms, Inc.}
  \country{USA}
}

\author{Rui Li}
\email{ruili23@meta.com}
\affiliation{%
  \institution{Meta Platforms, Inc.}
  \country{USA}
}

\author{Zihang Fu}
\email{fuzihang@meta.com}
\affiliation{%
  \institution{Meta Platforms, Inc.}
  \country{USA}
}

\author{Tongyi Tang}
\email{tongyitang@meta.com}
\affiliation{%
  \institution{Meta Platforms, Inc.}
  \country{USA}
}

\author{Zhengyu Zhang}
\email{zhengyuzhang@meta.com}
\affiliation{%
  \institution{Meta Platforms, Inc.}
  \country{USA}
}

\author{Wen-Yen Chen}
\email{wychen@meta.com}
\affiliation{%
  \institution{Meta Platforms, Inc.}
  \country{USA}
}

\author{Nima Noorshams}
\email{nshams@meta.com}
\affiliation{%
  \institution{Meta Platforms, Inc.}
  \country{USA}
}

\author{Nirav Jasapara}
\email{niravj@meta.com}
\affiliation{%
  \institution{Meta Platforms, Inc.}
  \country{USA}
}

\author{Xiaowen Ding}
\email{xwding@meta.com}
\affiliation{%
  \institution{Meta Platforms, Inc.}
  \country{USA}
}

\author{Ellie Wen}
\email{ellie.wen@meta.com}
\affiliation{%
  \institution{Meta Platforms, Inc.}
  \country{USA}
}

\author{Xue Feng}
\email{xfeng@meta.com}
\affiliation{%
  \institution{Meta Platforms, Inc.}
  \country{USA}}

\renewcommand{\shortauthors}{Zhang et al.}
\begin{abstract}
Optimizing conversions is crucial in modern online advertising systems, enabling advertisers to deliver relevant products to users and drive business outcomes. However, accurately predicting conversion events remains challenging due to variable time delays between user interactions (e.g., impressions or clicks) and the actual conversions. These delays vary substantially across advertisers and products, necessitating flexible optimization windows tailored to specific conversion behaviors. To address this, we propose a novel \textit{Personalized Interpolation} method that extends existing models based on fixed conversion windows to support flexible advertiser-specific optimization windows. Our method enables accurate conversion estimation across diverse delay distributions without increasing system complexity. We evaluate the effectiveness of the proposed approach through extensive experiments using a real-world ads conversion model. Our results show that this method achieves both high prediction accuracy and improved efficiency compared to existing solutions. This study demonstrates the potential of our Personalized Interpolation method to improve conversion optimization and support a wider range of advertising strategies in large-scale online advertising systems.
\end{abstract}



\received{16 May 2025}
\received[revised]{16 May 2025}
\received[accepted]{16 May 2025}

\maketitle


\section{Introduction}

Data freshness plays a pivotal role in the efficacy of online advertising recommendation systems \cite{labrinidis2003balancing, lee2013real}. The precision of predicting user conversions depends heavily on how promptly users’ latest interests are reflected in both labels and features. Over the past decades, the advertising industry has made substantial strides in increasing model training frequency, evolving from daily updates to hourly and even real-time processing. Despite these advancements, a significant challenge remains on the advertiser’s side, particularly regarding their preference to optimize for conversions that may occur long after a user is exposed to an ad. This duration, known as the \textit{optimization window}, is crucial as it determines the metrics by which ad performance is evaluated.

Long optimization windows pose a formidable challenge to ad delivery systems, since predictive models must endure lengthy delays to accumulate the complete label distributions needed to align with advertisers’ true objectives. Such delays can severely degrade performance; for example, a model that waits seven days to train on complete data may become obsolete for predicting conversions within that period. To address challenges associated with data delays, a new research domain — delayed feedback modeling (DFM) — has emerged \cite{chapelle2014modeling, ktena2019addressing, yasui2020feedback}. DFM employs Bayesian inference and innovative architectures to balance data freshness and completeness, enabling optimization for fixed long windows.

Despite advancements in managing fixed long optimization windows, significant gaps remain in accommodating advertisers’ precise needs. The length of the optimization window often varies depending on the nature of the advertiser’s products. For instance, gaming advertisers targeting new users may face longer cycles from app installation to specific in-app events, with this duration varying by game type and target audience. These advertisers often have deeper insights into appropriate conversion windows for their products. Similarly, e-commerce advertisers seeking to boost purchases may operate within different time frames influenced by promotional periods prior to counting conversions. These diverse requirements create unique opportunities for ad companies to develop machine learning solutions that align more closely with advertiser goals.

Ideally, recommendation systems should be versatile enough to support any optimization window specified by advertisers. However, our literature review reveals a \textit{lack of} prior research addressing this capability despite its practical importance. Developing a \textit{flexible optimization window (FOW)} ad delivery system introduces new challenges in both modeling and infrastructure. Beyond managing data delays, scalability becomes a bottleneck, requiring resource-efficient solutions capable of supporting varied window scenarios.


In this paper, we develop an efficient and scalable method for FOW estimation that requires minimal system resources. Our major contributions are as follows:
\begin{itemize}[leftmargin=*, itemindent=0pt]
    \item We conceptualize flexible optimization window estimation as approximating the cumulative distribution function (CDF) of conversion event times. Leveraging the monotonic increasing and concave properties of the CDF, we propose a novel personalized interpolation method to efficiently estimate conversions with flexible optimization windows. Our approach integrates seamlessly with existing production black-box models without incurring additional computational cost or system overhead.
    \item Our personalized interpolation method introduces a key parameter, the \textit{interpolation factor}, which governs how conversion estimates are adjusted across different optimization windows. Since this factor is not directly observable, we propose three distinct parametric functions to approximate it, each offering different trade-offs in terms of complexity, interpretability, and performance. To better understand their effectiveness, we conduct an ablation study to analyze and compare their performance.
    \item To evaluate the effectiveness of our proposed method, we perform extensive experiments comparing it against both the theoretical upper bound and practical baselines suitable for production. Results indicate that our method not only achieves efficient FOW estimation but also delivers prediction accuracy close to the theoretical optimum — without requiring additional training or infrastructure resources.
\end{itemize}
The rest of this paper is organized as follows. Section~\ref{Sec. Prelim} provides preliminaries and reviews related work. Section~\ref{Sec.Problem} formally formulates the FOW problem based on the long optimization window framework and highlights key challenges from modeling and system perspectives. Section~\ref{Sec. Method} details our proposed Personalized Interpolation method and its integration with production black-box models. Section~\ref{Sec. Exp} presents experimental results, and Section~\ref{Sec.Conclusion} concludes the paper.

\begin{figure}
    \centering
    \includegraphics[width=0.95\linewidth]{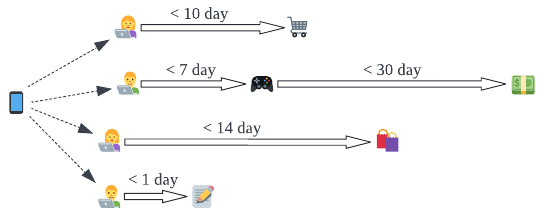}
    \vspace{-.1in}
    \caption{Illustration of varying conversion cycles from user interaction with ad impressions to final conversion events. The length of these cycles differs depending on the product type or conversion goal (e.g., form submission, game level achievement, or purchase). As a result, advertisers are motivated to optimize for different conversion windows tailored to their specific objectives.}
    \label{fig:multi_conv_cycl}
    \vspace{-.1in}
\end{figure}


\section{Related Works}\label{Sec. Prelim}

\noindent\textbf{Delayed Feedback Modeling for Conversion Prediction.} 
Modeling conversion delays has been extensively studied in online advertising. Early work focused on explicitly modeling the time between impressions or clicks and conversions. Chapelle (2014)~\cite{chapelle2014modeling} and Ji et al. (2016)~\cite{ji2016probabilistic} proposed parametric models using exponential and Weibull distributions, respectively, though these often fail to capture complex delay patterns due to strong distributional assumptions.

To improve flexibility, Yoshikawa et al. (2018)~\cite{yoshikawa2018nonparametric} applied kernel density estimation (KDE). More recently, Wang et al. (2020)~\cite{wang2020delayed} jointly modeled click-through rate (CTR), conversion rate (CVR), and conversion delay by discretizing conversion time and applying softmax classification.

Another research direction addresses delayed feedback via label correction and reweighting. FSIW~\cite{yasui2020feedback} uses importance sampling to adjust loss weights based on elapsed time. Extensions such as nnDF and DEFUSE~\cite{yasui2022learning, chen2022asymptotically} provide unbiased CVR estimation, while UCL~\cite{wang2023unbiased} jointly trains CVR and label correction models. ES-DFM~\cite{yang2021capturing} dynamically samples cut-off times to balance data freshness and label completeness. Several of these methods~\cite{ktena2019addressing, yang2021capturing} have shown success in industrial settings.

However, most approaches require dedicated delay models or complex training pipelines, increasing deployment costs. In contrast, our work reuses outputs from existing models to estimate conversion rates at arbitrary elapsed times—offering a more practical and efficient solution.
\smallskip

\noindent\textbf{Survival Analysis and Interpolation Method.} 
FOW estimation can be framed as approximating the cumulative distribution function (CDF) of conversion time, similar to problems in survival analysis~\cite{jenkins2005survival, elandt2014survival, barbieri2016improving, wang2019machine}. Survival models, widely used in healthcare and engineering, estimate the survival function $S(t) = \mathbb{P}(T > t)$, where $T$ denotes event time.

Classical approaches include the Kaplan–Meier estimator~\cite{kaplan1958nonparametric}, Cox proportional hazards model~\cite{cox1972regression, breslow1975analysis}, and accelerated failure time model~\cite{wei1992accelerated}. Recent deep learning variants such as DeepSurv~\cite{katzman2018deepsurv} and DASA~\cite{nezhad2019deep} improve performance but require intensive computation and model design, limiting industrial scalability.

Interpolation techniques, widely used in statistics and numerical analysis~\cite{de1978practical, gautschi2011numerical}, offer an efficient alternative for estimating values between known data points. Prior works~\cite{whittemore1986survival, gray1992flexible, molinari2001regression} have combined interpolation with survival models to handle large or incomplete datasets.

Building on this idea, our approach uses interpolation over predictions from existing models to efficiently estimate conversion rates for arbitrary time windows—offering both accuracy and ease of deployment.

\section{Problem Statement}\label{Sec.Problem}

In this section, we first describe the ads conversion model designed for long optimization window scenarios. Building on this formulation, we then extend the framework to address the Flexible Optimization Window (FOW) problem, and discuss the key challenges associated with enabling flexibility in both modeling and system implementation.

\subsection{Preliminary for Long Optimization Window}

We define the probability of an ad conversion event $e$ occurring within a given optimization window $T$ as $\mathbb{P}(\tau \in (0, T] \mid e)$, where $\tau$ denotes the time elapsed from a user's click on the ad to the actual conversion. The value of $T$ represents the optimization window, i.e., the maximum delay allowed for a conversion to be attributed to the ad click.


For short optimization windows (e.g., $T \leq 1$ day), the model can typically be trained directly, as the performance is relatively robust to short delays in label availability. However, when the optimization window becomes long (e.g., $T = 7$ days), delayed feedback becomes a critical bottleneck, significantly impacting model training and prediction accuracy.

A common approach to mitigate this issue is to decompose the long optimization window conversion probability into more manageable sub-problems - particularly by leveraging predictions from a short window where data is fresher and more accurate. Specifically, the conversion probability over a long window can be decomposed as follows:
\begin{align}
    &\mathbb{P}(\tau \in (0, T_l]|e) = \mathbb{P}(\tau \in (0, T_s]|e) + \mathbb{P}(\tau \in (T_s, T_l]|e) \label{Eq.LOW-1}
    \\
    &\!=\! \mathbb{P}(\tau\! \in\! (0, T_s]|e) \!+\! \Big(\!1 \!- \!\mathbb{P}(\tau \!\in\! (0, T_s]|e)\! \Big) \!\cdot \!
    \mathbb{P}(\tau \!\in\! (T_s, T_l] | \tau > T_s,e) \label{Eq.LOW-2}
\end{align}
Here, $T_s$ and $T_l$ denote the short and long optimization windows, respectively. This formulation decomposes the long window prediction into two components: the probability of conversion within the short window $(0, T_s]$, and the conditional probability of converting in the remaining window $(T_s, T_l]$, given that the conversion did not occur before $T_s$.
Equations~\eqref{Eq.LOW-1} and~\eqref{Eq.LOW-2} represent two alternative decompositions. Notably, \eqref{Eq.LOW-2} explicitly conditions the second term on the event $\tau > T_s$, thereby making more effective use of the short-window prediction $\mathbb{P}(\tau \in (0, T_s] \mid e)$, which is typically more accurate due to fresher data availability.

\subsection{Problem Formulation of FOW and its Challenges}

Building on the long optimization window framework, we can reuse the formulation from \eqref{Eq.LOW-1} or \eqref{Eq.LOW-2} by replacing the long window $T_l$ with the desired window $T_f$. While the short-window prediction $\mathbb{P}(\tau \in (0, T_s] \mid e)$ can still be leveraged, we now require an additional model to estimate either $\mathbb{P}(\tau \in (T_s, T_f] \mid e)$ or the conditional probability $\mathbb{P}(\tau \in (T_s, T_f] \mid \tau > T_s, e)$.

However, this approach quickly becomes impractical when scaling to support multiple advertisers with varying optimization windows. For example, supporting daily windows from day 1 to day 7 would necessitate training at least seven distinct conversion models. This does not even account for scenarios beyond 7 days or finer-grained intervals such as hourly windows. Such a solution would lead to significantly increased training costs, high maintenance overhead, and greater risks to system stability.

Additional challenges arise in the production environment, including:  
1) increased complexity in serving logic, as each optimization window would require distinct configurations to route requests to the correct models;  
2) the need to build and maintain multiple data pipelines for label generation across different window lengths, which would impose high demands on both storage and computations.

Given these challenges, it is crucial to develop resource-efficient and scalable solutions that can flexibly support a wide range of optimization windows without incurring substantial system overhead.


\section{Methodology}\label{Sec. Method}

\begin{figure}
    \centering
    \includegraphics[width=0.8\linewidth]{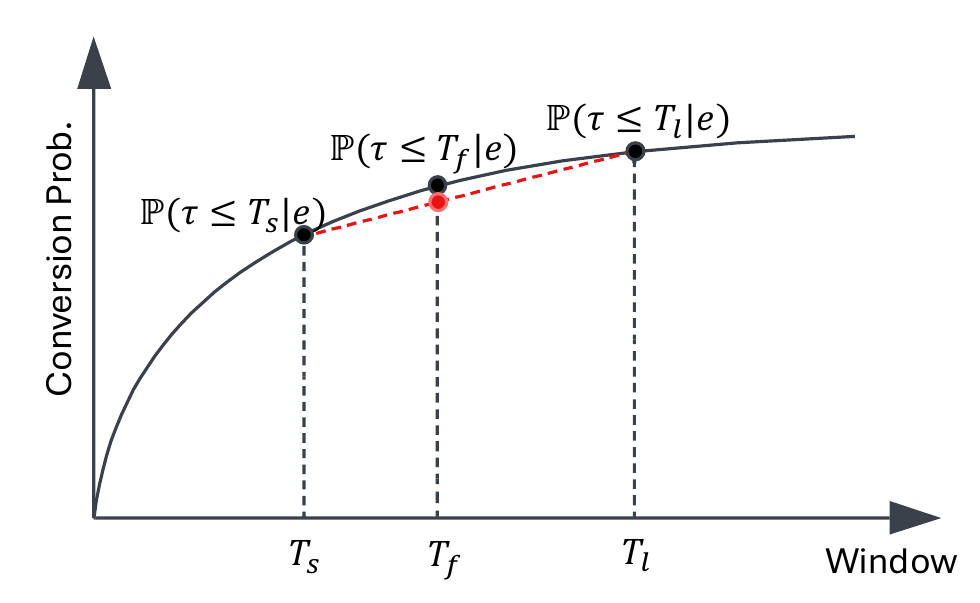}
    \vspace{-.1in}
    \caption{
    Illustration of the interpolation framework on the conversion CDF. Given the conversion probabilities $\mathbb{P}(\tau \le T_s|e)$ and $\mathbb{P}(\tau \le T_l|e)$, the probability at an intermediate window $\mathbb{P}(\tau \le T_f|e)$ (black dot) can be approximated via a linear combination of the two endpoints (red dot).}
    \label{fig:ill_interpolation}
    \vspace{-.1in}
\end{figure}

 \begin{figure*}
\centering
    \begin{tabular}{ccc}
        {\includegraphics[width=0.25\linewidth]{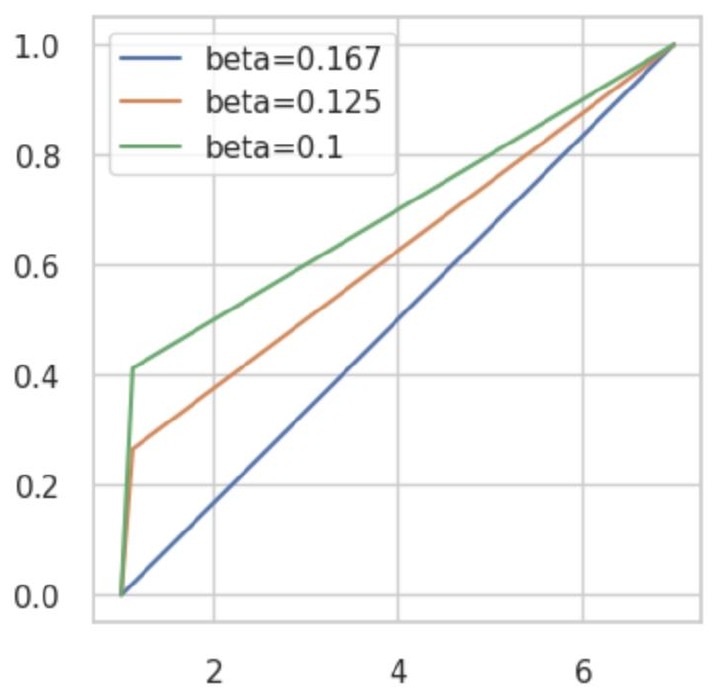}} & {\includegraphics[width=0.25\linewidth]{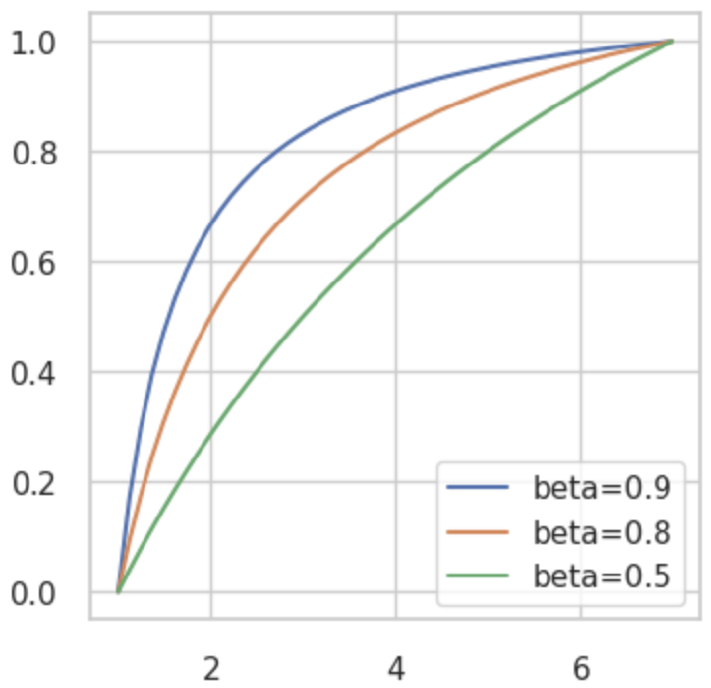}} & {\includegraphics[width=0.25\linewidth]{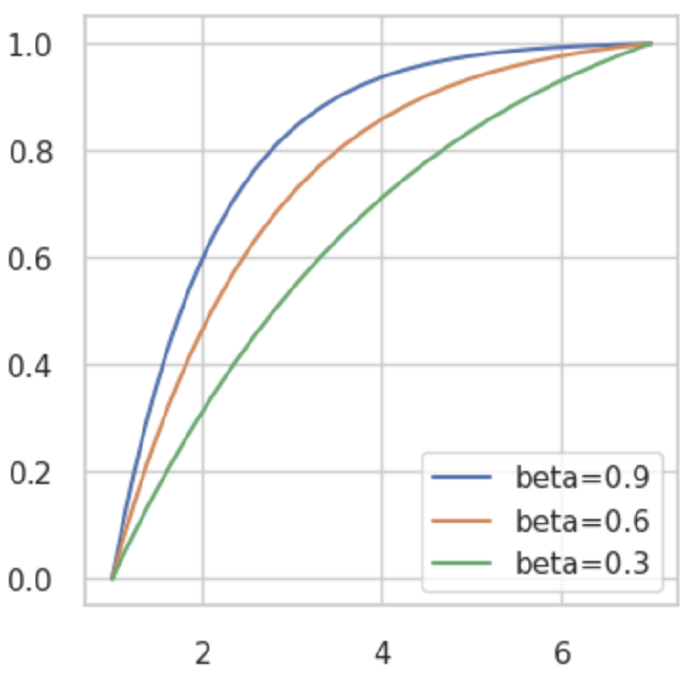}}\\
        \small{(a) Linear Function} &  \small{(b) Rational Function} & \small{(c) Exponential Function}\\
    \end{tabular}
    \vspace{-.1in}
    \caption{
    Illustration and comparison of the three proposed interpolation functions under the setting $T_s = 1$ and $T_l = 7$.
    }
    \label{fig:illustration}
    \vspace{-.1in}
\end{figure*}

In this section, we first introduce the framework of the proposed \textit{Personalized Interpolation} method and provide the underlying rationale behind its design. 
Next, in Section~\ref{Sec. Design of INTP}, we present several implementation variants of the interpolation mechanism and analyze their respective trade-offs. 
Finally, we describe how to seamlessly integrate the proposed method with existing black-box conversion prediction models in production environments.

\subsection{Framework of Personalized Interpolation}

In our approach, we assume the availability of conversion probability predictions for two fixed optimization windows: $\mathbb{P}(\tau \le T_s|e)$ and $\mathbb{P}(\tau \le T_l|e)$, where $T_s$ and $T_l$ represent short and long optimization windows, respectively. Our goal is to estimate the conversion probability for an intermediate window $T_f \in (T_s, T_l)$.

Treating the conversion time $\tau$ as a random variable, the probability of conversion within any given window $T$ corresponds to the cumulative distribution function (CDF), i.e., $\mathbb{P}(\tau \le T|e)$. Since the CDF is a monotonically increasing function (see Figure~\ref{fig:ill_interpolation}), the conversion probability at window $T_f$ can be approximated using a weighted combination of the two known predictions as follows:
\begin{align}\label{Eq. linear_combination}
    \mathbb{P}(\tau \le  T_f|e) = (1 -  \alpha(T_s,T_f,T_l) ) \cdot&\mathbb{P}(\tau\le  T_s|e) \notag\\
    &+ \alpha(T_s,T_f,T_l)\cdot \mathbb{P}(\tau \le T_l|e).
\end{align}

In Eq.~\eqref{Eq. linear_combination}, the interpolation coefficient $\alpha(T_s,T_f,T_l)$ corresponds to the normalized mass of the CDF between $T_s$ and $T_f$:
\vspace{-0.02in}
\[
\alpha(T_s,T_f,T_l) = \frac{\mathbb{P}(\tau \in (T_s,T_f]|e)}{\mathbb{P}(\tau \in (T_s,T_l]|e)} \in [0,1].
\vspace{-0.02in}
\]
However, the exact value of $\alpha(T_s,T_f,T_l)$ is typically unknown, as it depends on the underlying distribution of $\tau$. Therefore, we treat $\alpha$ as a hyperparameter and interpret Eq.~\eqref{Eq. linear_combination} as a general interpolation scheme that estimates $\mathbb{P}(\tau \le T_f|e)$ using only the predictions at $T_s$ and $T_l$.
Since this interpolation is performed at the individual event level, we refer to this approach as \textit{Personalized Interpolation}.
\begin{figure*}
    \centering
    \includegraphics[width=0.9\linewidth]{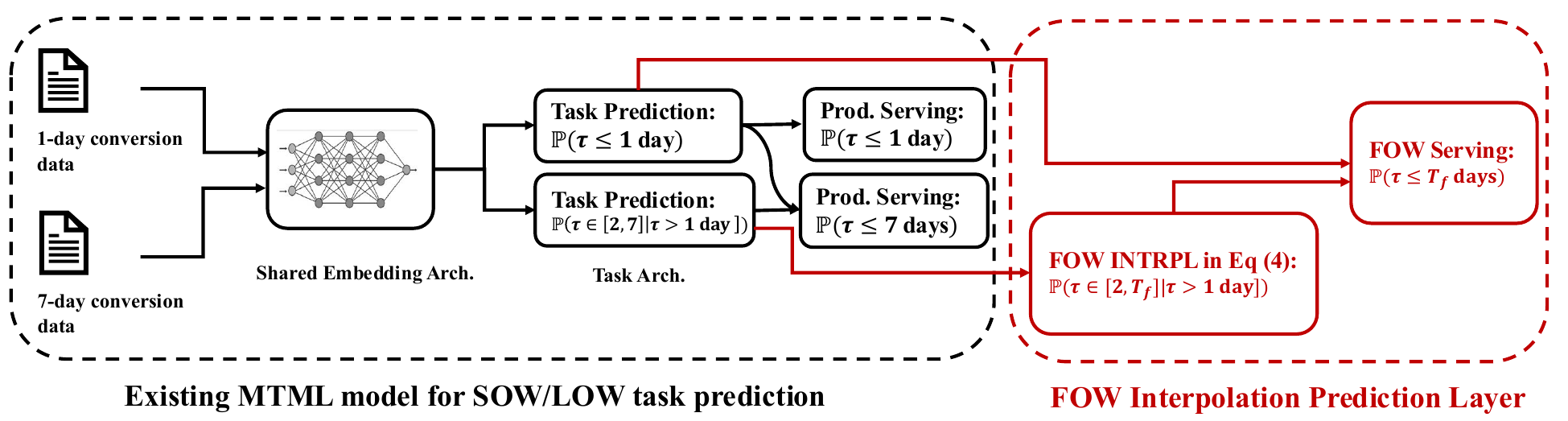}
    \vspace{-.1in}
    \caption{
    Model and system architecture of the SOW/LOW multi-task multi-label framework integrated with the flexible optimization window (FOW) interpolation layer.
    }
    \label{fig:model_arch}
    \vspace{-.1in}
\end{figure*}

Furthermore, when the conditional probability $\mathbb{P}(\tau \in (T_s,T_l] \mid \tau >T_s)$ is of interest, the following proposition shows that, conditioning on no conversion occurring within the initial window $T_s$, the probability of conversion within the target window $T_f$ is simply a scaled version of the probability within $T_l$:
\begin{proposition}\label{Prop.Cond}
Assume predictions are available for two optimization windows $T_s$ and $T_l$, and we seek to estimate the conversion probability for a new window $T_f \in [T_s, T_l]$. Then, the interpolation framework in Eq.~\eqref{Eq. linear_combination} implies:
\begin{align}\label{Eq.conditional}
    \mathbb{P}(\tau\in(T_s, T_f]~|~\tau >T_s) = \alpha(T_s,T_f,T_l) \cdot \mathbb{P}(\tau \in (T_s,T_l]~|~\tau >T_s),
\end{align}
where $\mathbb{P}(\tau \in (T_s, T_f] \mid \tau >T_s)$ denotes the conditional probability of a conversion occurring in the interval $(T_s, T_f]$ given that no conversion occurred within $T_s$, and similarly for $\mathbb{P}(\tau \in (T_s, T_l] \mid \tau >T_s)$.
\end{proposition}
Proposition~\ref{Prop.Cond} suggests we can use a single scalar to interpolate conversions from one optimization window to another. This result reinforces the central role of the interpolation factor $\alpha(T_s, T_f, T_l)$ in our personalized interpolation framework. For notational simplicity, we will henceforth refer to $\alpha(T_s, T_f, T_l)$ simply as $\alpha$.

\subsection{Design of Interpolations Factor}\label{Sec. Design of INTP}

With the framework in Eq.~\eqref{Eq. linear_combination}, we now proceed to the design of the unknown interpolation factor $\alpha(T_s,T_f,T_l)$. In our work, we propose three types of parameterized functions: linear, rational, and exponential. Below, we describe each design in detail and compare them in Figure~\ref{fig:ill_interpolation} using the setting $T_s=1$ and $T_l=7$:
\begin{itemize}[leftmargin=15pt, itemindent=0pt]
  \item[1).] \textit{Linear function interpolation}: The simplest design adopts a linear form, where the interpolation factor is given by $\alpha(T_s,T_f,T_l) = \beta\cdot (T_f-T_l) + 1$, and $\beta$ controls the slope (Fig.~\ref{fig:illustration}(a)). According to Eq.~\eqref{Eq.conditional}, this implies that $\mathbb{P}(\tau \in (T_s, T_f] | \tau >T_s)$ grows linearly toward $\mathbb{P}(\tau \in (T_s, T_l] | \tau >T_s)$. This design is suitable when the CDF exhibits a relatively flat curvature. To ensure $\alpha \in [0,1]$, the feasible range for $\beta$ is $[0,1/(T_l-T_s)]$. However, when $\beta \neq 1/(T_l-T_s)$, the condition $\alpha(T_s,T_s,T_l)=0$ introduces an abrupt drop near the boundary point $T_s$.
  
  \item[2).] \textit{Rational function interpolation}: This design introduces curvature into the interpolation using a rational function: $\alpha(T_s,T_f,T_l) = \frac{T_f-T_s}{\beta\cdot(T_f-T_l)+T_l-T_s}$, where $\beta$ is a hyperparameter controlling the concavity (Fig.~\ref{fig:illustration}(b)). Unlike the linear case, this formulation ensures a smooth transition without abrupt changes at the boundary, making it suitable for more curved CDFs.
  
  \item[3).] \textit{Exponential function interpolation}: Assuming the conversion time $\tau$ follows a zero-inflated exponential distribution $\mathbb{P}(\tau \le T_f|e) = p_{\text{conv}}(e) \cdot (1 - \exp(-\beta \cdot T_f))$, where $p_{\text{conv}}(e)=\mathbb{P}(\tau \le \infty|e)$ is the ultimate conversion probability and $\beta \ge 0$, we can derive the interpolation factor as:
  \vspace{-0.05in}
  \[
  \alpha(T_s,T_f,T_l) = \frac{\exp(-\beta\cdot T_s) - \exp(-\beta\cdot T_f)}{\exp(-\beta\cdot T_s) - \exp(-\beta\cdot T_l)}.
  \vspace{-0.05in}
  \]
  As illustrated in Fig.~\ref{fig:illustration}(c), this design yields a smooth, concave interpolation curve, aligning well with scenarios where conversion likelihood decays exponentially over time.
\end{itemize}
In the above three designs of the interpolation factor, we introduce a hyperparameter $\beta$ to control the shape of each interpolation function. Note that while we use the same symbol $\beta$ across all designs, its meaning and effect differ depending on the specific formulation. In Section~\ref{sec.ablation_study}, we will conduct an ablation study to investigate the impact of $\beta$ on estimation performance.

\subsection{Integration with Black-Box Models}



We now describe how the proposed method can be seamlessly integrated with black-box conversion prediction models in production environments.
In our interpolation framework (Eq.~\eqref{Eq. linear_combination}), the conversion probabilities for the two optimization windows $\mathbb{P}(\tau \le T_s|e)$ and $\mathbb{P}(\tau \le T_l|e)$ are required. In practice, these can be obtained using either two separate deep neural network models or a single multi-task model, resulting in predictions denoted by $\widehat{\mathbb{P}}(\tau \le T_s|e) = f_{T_s}(e)$ and $\widehat{\mathbb{P}}(\tau \le T_l|e) = f_{T_l}(e)$.

Once the models are trained, an interpolation layer can be added on top of the outputs at inference time to compute the desired prediction for a flexible window $T_f$:
\vspace{-.05in}
\[
\widehat{\mathbb{P}}(\tau \le T_f|e) = (1 - \alpha) \cdot f_{T_s}(e) + \alpha \cdot f_{T_l}(e),
\vspace{-.05in}
\]
where $\alpha$ is chosen according to a predefined interpolation design. This interpolation step is applied only during the evaluation stage and incurs no additional training or modeling cost.

Similarly, our approach can be applied in the conditional probability setting using Proposition~\ref{Prop.Cond}. Assume the models can provide estimates of $\widehat{\mathbb{P}}(\tau \le T_s|e) = f_{T_s}(e)$ and $\widehat{\mathbb{P}}(\tau \in (T_s, T_l] \mid \tau > T_s) = f_{T_l|T_s}(e)$. Then, by applying the law of total probability, we compute the flexible-window prediction as:
\vspace{-.05in}
\[
\widehat{\mathbb{P}}(\tau \le T_f|e) = f_{T_s}(e) + (1 - f_{T_s}(e)) \cdot (\alpha \cdot f_{T_l|T_s}(e)).
\vspace{-.05in}
\]
We adopt this conditional probability formulation in our experiments and provide implementation details in Section~\ref{Sec. Exp}.


\section{Experiment}\label{Sec. Exp}


\subsection{Experiment Setup}

We conducted experiments using an ads conversion model based on a multi-task multi-label (MTML) consolidation framework~\cite{huang2013multi, zhang2018overview, ma2022online}, designed to predict conversions under two optimization windows: a short optimization window (SOW) of 1 day and a long optimization window (LOW) of 7 days. The overall model structure is illustrated in Fig.~\ref{fig:model_arch}.
The model takes merged conversion data from both the 1-day (SOW) and 7-day (LOW) windows as input, and outputs two predictions:
\begin{itemize}
    \item The 1-day conversion probability, $\widehat{\mathbb{P}}(\tau \le \text{1 day})$;
    \item The conditional probability of conversion in days 2 to 7, given no conversion in the first day, $\widehat{\mathbb{P}}(\tau \in \text{[2, 7]}~|~\tau > \text{1 day})$.
\end{itemize}
Using these two outputs, the overall 7-day conversion probability can be recovered by:
$\widehat{\mathbb{P}}(\tau \le \text{7 days}) = \widehat{\mathbb{P}}(\tau \le \text{1 day}) + (1 - \widehat{\mathbb{P}}(\tau \le \text{1 day})) \cdot \widehat{\mathbb{P}}(\tau \in \text{[2, 7]}~|~\tau > \text{1 day}).$
This MTML consolidation approach provides an efficient solution to the long optimization window problem. In our experiments, we use the existing MTML model to estimate conversion probabilities for flexible optimization windows (FOW) with $T_f \in (1, 7)$ using the proposed interpolation framework.


\begin{figure}
    \centering
    \includegraphics[width=1.0\linewidth]{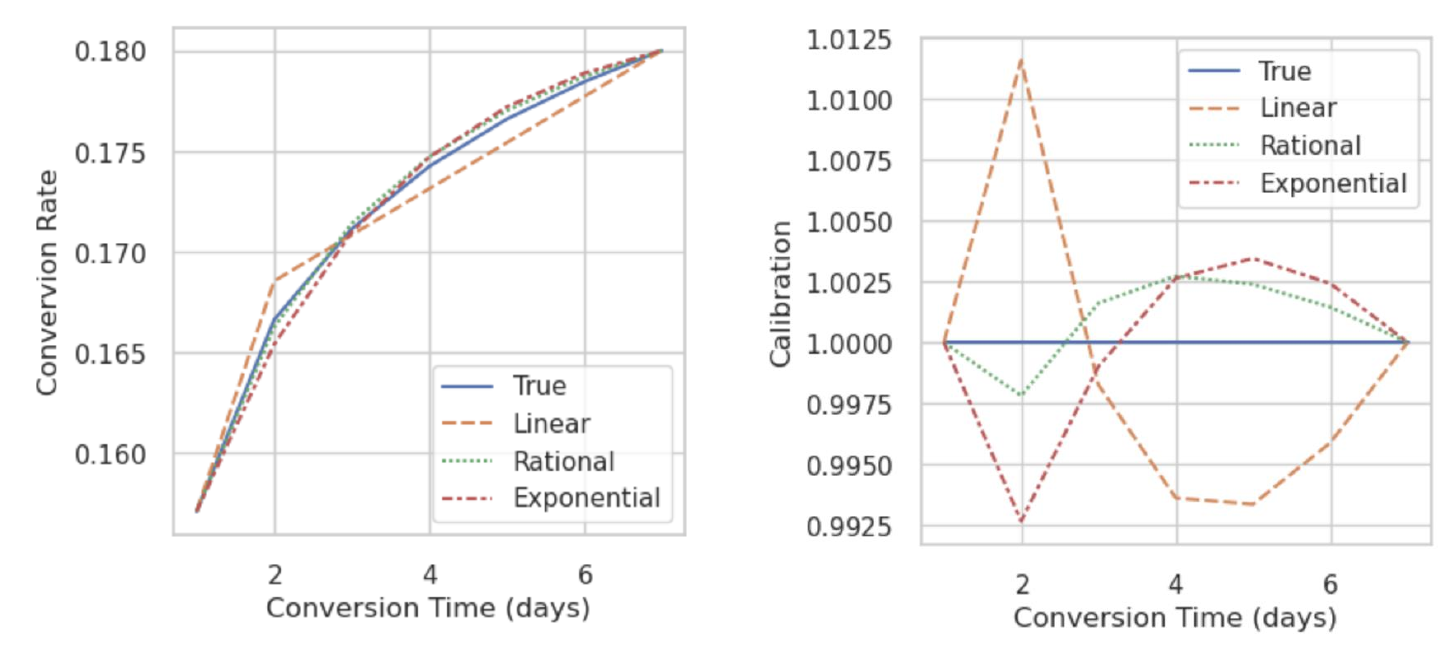}
    \vspace{-.23in}
    \caption{Conversion data analysis on real impression data collected from 2023-12-27 to 2024-02-04. The left panel shows the average conversion rates with three different interpolation estimations, while the right panel presents the corresponding calibration scores. For all three interpolation methods, the hyper-parameter $\beta$ values are set as listed in Table~\ref{tab:beta_setup}.}
    \label{fig:real_data}
    \vspace{-.1in}
\end{figure}

\begin{table}
    \centering
    \caption{Hyper-parameter $\beta$ values for interpolation methods in the experiments discussed in Sec.~\ref{Sec.comparison}.}
    \vspace{-.1in}
    \begin{tabular}{c|c|c|c}
    \hline
         Type & Lienar & Rational & Exponential \\
         \hline
        $\beta$ value & 0.1 & 0.7 & 0.4\\
    \hline
    \end{tabular}
    \label{tab:beta_setup}
    \vspace{-.2in}
\end{table}

In our experiments, the training data is the impression data collected from ads delivery system over 40 days. Before model training and evaluation, we performed an analysis of the conversion time distribution. 
The left panel of Fig.~\ref{fig:real_data} depicts the true conversion probability curve (i.e., the CDF of $\tau$) as the solid blue line. The curve exhibits a concave shape, increasing from $\mathbb{P}(\tau \le \text{1 day}) = 0.157$ to $\mathbb{P}(\tau \le \text{7 days}) = 0.180$. Using these true values at the 1-day and 7-day windows, we apply the proposed interpolation methods and plot the estimated conversion probabilities as dashed curves. These interpolated curves closely follow the true conversion curve.
To quantitatively assess the difference between the true and interpolated conversion rates, we show calibration scores in the right panel of Fig.~\ref{fig:real_data}. The calibration score~\cite{he2014practical} is defined as the ratio between the interpolated estimation, $\widehat{\mathbb{P}}_{\text{INTPRL}}(\tau \le T_f \text{ day})$, and the true value, $\mathbb{P}(\tau \le T_f \text{ day})$, for $T_f \in [1,7]$. The results indicate that all three interpolation methods yield calibration scores close to 1, with the rational function interpolation exhibiting the smallest deviation.
Based on this conversion data analysis, we are confident that our proposed interpolation methods provide an accurate and reliable fit for flexible optimization window (FOW) prediction.

\subsection{Performance Comparison}

\subsubsection{Comparison with Existing methods}\label{Sec.comparison}

Now, we proceed to evaluate our proposed interpolation methods by comparing them against five competitive baselines:
\begin{itemize}[leftmargin=*, itemindent=0pt]
    \item \textit{$\mathbb{P}_\text{1d}$-Model}: This method directly uses the SOW prediction $\mathbb{P}(\tau \!\le\! \text{1 day})$ as an estimate for $\mathbb{P}(\tau \!\le\! T_f~\text{day})$ for all $T_f \!\in\! [1,7]$. Since it ignores potential conversions beyond 1 day, it tends to be under-calibrated (i.e., calibration scores below 1).
    
    \item \textit{$\mathbb{P}_\text{7d}$-Model}: This method treats the LOW prediction $\mathbb{P}(\tau \le \text{7 day})$ as a proxy for $\mathbb{P}(\tau \le T_f~\text{day})$ for $T_f \in [1,7]$. It overestimates early conversions, resulting in over-calibration (i.e., calibration scores above 1).
    
    \item \textit{7-TaskHead Model (7THM)}: A newly designed multi-task model with 7 separate heads, each trained to predict conversion by day $T_f$ for $T_f = 1$ to $7$. The model’s effectiveness depends on proper task balancing and sufficient training data for each head.
    
    \item \textit{Dedicated Models (DMs)}: This approach trains 6 individual multi-task models. Each model predicts both $\mathbb{P}(\tau \le \text{1 day})$ and the conditional probability $\mathbb{P}(\tau \in T_f~\text{to 7 days} \mid \tau > \text{1 day})$ for a specific $T_f \in [2,7]$, which are composite into $\mathbb{P}(\tau \le \text{7 day})$. However, the training labels for these models are delayed, reflecting real production constraints.
    
    \item \textit{No Delay Upper Bound (NDUB)}: Identical to DMs in structure, but trained with up-to-date (non-delayed) conversion labels. This method is not feasible in production but serves as a theoretical upper bound on achievable performance.
\end{itemize}
For our interpolation-based methods, we use the hyper-parameter values in Table~\ref{tab:beta_setup}, selected through simple tuning. An ablation study analyzing the impact of $\beta$ is provided in Section~\ref{sec.ablation_study}.

\begin{figure}
    \centering
    \includegraphics[width=0.9\linewidth]{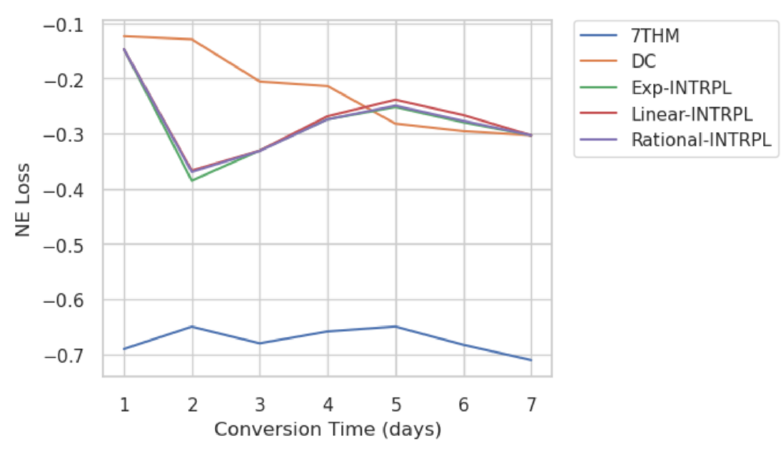}
    \vspace{-.1in}
       \caption{Normalized entropy (NE) loss (\%) compared with NDUB. The less negative value means better performance.}
    \label{fig:NDUB}
    \vspace{-.1in}
\end{figure}

\begin{figure*}
    \centering
    \includegraphics[width=0.95\linewidth]{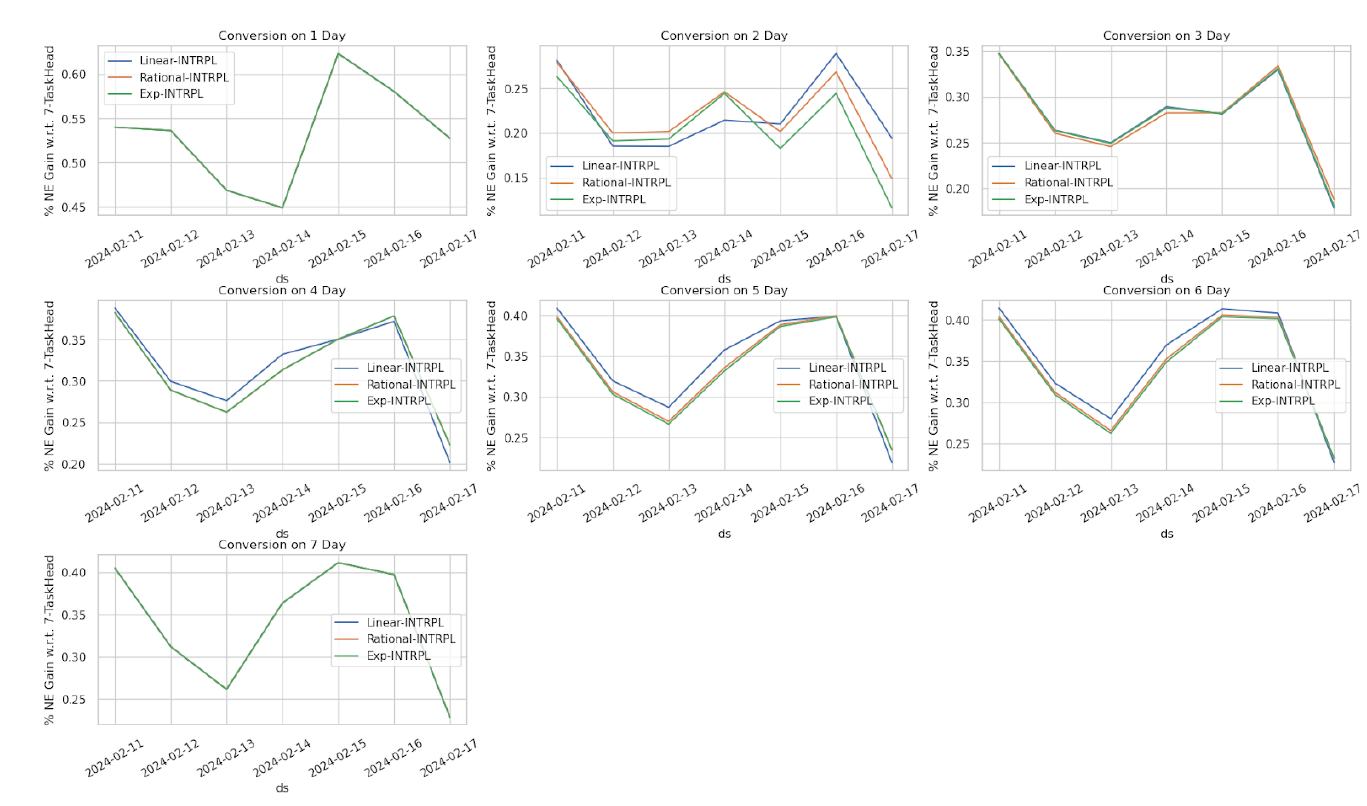}
    \captionsetup{margin={200pt,0pt},skip=-70pt}
  \caption{Performance comparison of recurring training (last for 1 week during 2024-02-11 and 2024-02-17) between our proposed interpolation methods and the 7THM model. It sets the 7THM model as the baseline and the NE improvement (\%) as the metric. The larger value of the NE improvement means the better performance.\\ \\ }
    \label{fig:recurring}
\vspace{-.2in}
\end{figure*}

\noindent\textbf{Experimental Results.} We evaluate prediction accuracy using the normalized entropy (NE) metric~\cite{he2014practical}, where lower NE values indicate better performance. The NE results for all methods are reported in Table~\ref{fig:exp_result}.
Our three interpolation methods achieve NE scores that closely match those of the Dedicated Models (DMs) baseline, indicating that the proposed interpolation framework is highly effective for production deployment. Unlike DMs, which requires training six separate models, our approach only requires a single light-weight model, substantially reducing both training and infrastructure costs.
The method using $\mathbb{P}_\text{1d}$ model yields the worst NE performance for estimating conversions beyond the first day, due to its under-calibrated nature. Conversely, the $\mathbb{P}_\text{7d}$ model approximation performs poorly in the early part of the conversion window (first three days) but improves as the prediction horizon lengthens. These observations align with the trend in Fig.~\ref{fig:real_data}, where the empirical conversion CDF is concave and flattens over time.
The 7-TaskHead Model (7THM) produces moderately competitive NE values, though consistently worse than those of our interpolation framework. We attribute this to the difficulty of maintaining balance across multiple tasks, which can hinder the performance improvements of such multi-head architectures.

To further analyze the performance gap between our proposed method and the theoretical upper bound, we use the results of the No Delay Upper Bound (NDUB) method as a baseline and compute the NE loss (\%) for the Dedicated Models (DMs) method, the 7-TaskHead Model (7THM), and our three interpolation methods. The results are visualized in Figure~\ref{fig:NDUB}.
The DC method consistently delivers the best performance, with NE loss increasing slightly from $-0.1\%$ to $-0.3\%$ as the conversion window extends from 1 day to 7 days. Our interpolation methods show an NE improvement of approximately $-0.3\%$ for conversion windows beyond 1 day. Interestingly, as the prediction horizon increases, our methods begin to outperform the DMs approach, highlighting their strengths in both prediction accuracy and training efficiency.
It is important to note that this training efficiency arises from the fact that the DMs method requires training six separate models, whereas our interpolation framework only requires a single model.
In contrast, the 7THM approach exhibits significantly larger NE losses, around $-0.7\%$, indicating suboptimal performance. Overall, these comparative results demonstrate that our proposed interpolation framework achieves a favorable trade-off between predictive accuracy and computational efficiency, making it particularly well-suited for large-scale deployment.


\begin{table*}[h]
\caption{Performance comparison of different methods for flexible optimization window (FOW) prediction. The table reports the normalized entropy (NE) values across various conversion windows. Lower NE values indicate better predictive performance.}
\vspace{-.1in}
\begin{tabular}{lrrrrrrrr}
\hline
\multirow{2}{*}{Conv. Day} & \multicolumn{8}{c}{Methods} \\ 
  & NDUB & DMs & \textbf{Linear-INTRPL} & \textbf{Rational-INTRPL} & \textbf{Exp-INTRPL} & 7THM & $\mathbb{P}_\text{1d}$ & $\mathbb{P}_\text{7d}$ \\
  \hline
1 & 0.5995 & 0.6002 & 0.6004 & 0.6004 & 0.6004 & 0.6036 & 0.6004 & 0.6159 \\
2 & 0.6051 & 0.6059 & 0.6073 & 0.6073 & 0.6074 & 0.6090 & 0.6177 & 0.6115 \\
3 & 0.6082 & 0.6095 & 0.6103 & 0.6103 & 0.6103 & 0.6124 & 0.6238 & 0.6123 \\
4 & 0.6110 & 0.6123 & 0.6126 & 0.6127 & 0.6127 & 0.6150 & 0.6287 & 0.6137 \\
5 & 0.6130 & 0.6148 & 0.6145 & 0.6146 & 0.6146 & 0.6170 & 0.6326 & 0.6150 \\
6 & 0.6144 & 0.6162 & 0.6161 & 0.6161 & 0.6161 & 0.6186 & 0.6360 & 0.6163 \\
7 & 0.6156 & 0.6175 & 0.6175 & 0.6175 & 0.6175 & 0.6200 & 0.6391 & 0.6175 \\
\hline
\end{tabular}\label{fig:exp_result}
\end{table*}

\subsubsection{Performance Check with Recurring Training}


To evaluate the long-term effectiveness of our methods, we conducted recurring training and evaluation over a one-week period, spanning from 2024-02-11 to 2024-02-17. Each day's model was trained using the most recent data and evaluated on the following week.
In this analysis, we use the 7-TaskHead model as a baseline and compare it against our personalized interpolation methods. Fig.~\ref{fig:recurring} presents the normalized entropy (NE) gains (\%) achieved by our methods relative to the baseline. The results show that our interpolation methods consistently outperform the 7-TaskHead model, achieving over 0.2\% NE gain across the evaluation period.
Notably, the NE differences among the three interpolation variants are minimal, suggesting all designs offer stable and robust performance. Additionally, all three interpolation methods yield identical NE values at the 1-day and 7-day boundaries, since their interpolated estimates align exactly with the original predictions at these endpoints.
\begin{figure}
    \centering
    \includegraphics[width=1\linewidth]{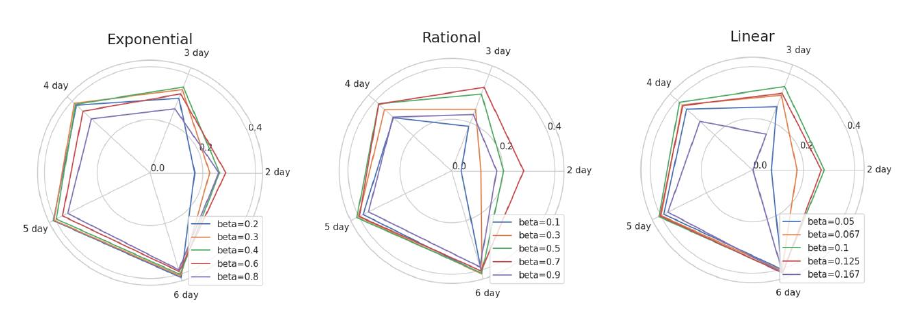}
    \caption{Ablation study on the hyper-parameter $\beta$ in our interpolation methods. The 7-TaskHead model serves as the baseline, and the NE improvement (\%) is reported as the evaluation metric. A larger shaded area indicates better overall accuracy and robustness to hyper-parameter variation.}
    \label{fig:ablation}
    \vspace{-.2in}
\end{figure}

\subsubsection{Ablation Study on Hyperparameters}\label{sec.ablation_study}

In the final part of our experiments, we conducted an ablation study to investigate the influence of the hyper-parameter $\beta$ on the accuracy and stability of our interpolation methods. Since $\beta$ is the only tunable parameter in our framework, we systematically varied its value and measured the corresponding NE gain (\%) over the 7-TaskHead baseline.
The results, shown in Figure~\ref{fig:ablation}, reveal that the exponential interpolation method is relatively less sensitive to changes in $\beta$, demonstrating greater robustness compared to the linear and rational function interpolations. Furthermore, for all interpolation types, the NE improvement diminishes as the prediction window approaches 1 day, which aligns with the reduced role of interpolation near the lower boundary.


\section{Conclusion}\label{Sec.Conclusion}


In this paper, we proposed a novel framework, termed \textit{personalized interpolation}, to effectively address the Flexible Optimization Window (FOW) problem in advertising conversion modeling. The framework leverages two conversion predictions corresponding to short and long optimization windows and interpolates between them to estimate conversion probabilities for arbitrary windows, thereby removing the need for direct supervision.
Our method is model-agnostic and can be seamlessly integrated with existing black-box conversion prediction systems commonly used in industrial environments, with negligible extra computational or training overhead.
Extensive experiments demonstrate that our interpolation-based approach consistently achieves competitive or superior predictive performance compared to alternative production-ready solutions. These results validate the effectiveness and practicality of the proposed method for real-world deployment in advertising systems.



\bibliographystyle{ACM-Reference-Format}
\bibliography{sample-base}

\end{document}